\documentclass[conference]{IEEEtran}

\usepackage{cite}
\usepackage{url}
\usepackage{amsmath,amssymb,amsfonts}
\usepackage{algorithm}
\usepackage[noend]{algpseudocode}
\usepackage{graphicx}
\usepackage[caption=false,font=footnotesize]{subfig}
\usepackage{textcomp}
\usepackage{array}
\usepackage{csquotes}
\usepackage{multirow}
\usepackage{xcolor}
\usepackage{makecell}
\usepackage{subcaption}
\usepackage[hidelinks]{hyperref}
\usepackage[font=footnotesize]{caption}
\usepackage{stfloats}
\usepackage{fancyhdr}
\usepackage{lettrine}
\usepackage{booktabs}
\definecolor{my_blue}{RGB}{0,0,255}

\newcolumntype{C}[1]{>{\centering\arraybackslash}p{#1}}

\begin{document}

\title{UNIFY: Unified Learning with One Model via Batch Norm Switching}

\author{
\IEEEauthorblockN{Shaharyar Ahmed Khan Tareen}
\textit{University of Houston}\\
Houston, TX, USA \\
stareen@cougarnet.uh.edu
\and
\IEEEauthorblockN{Xiaojing Yuan}
\textit{University of Houston} \\
Houston, TX, USA \\
xyuan@uh.edu
}

\maketitle


\begin{abstract}
Neural networks are typically trained separately for individual datasets or tasks, resulting in many task-specific models. This redundancy increases memory footprint, deployment costs, and energy consumption limiting their scalability. We propose UNIFY, a unified learning framework that enables a single model to jointly learn and perform inference across multiple datasets via Batch Normalization Switching (BNS). Instead of maintaining separate models for each dataset, UNIFY shares the parameter-intensive backbone while maintaining dataset-specific BN statistics and parameters, along with dataset-specific classification heads. We validate our framework using five datasets (MNIST, Fashion-MNIST, CIFAR-10, CIFAR-100, Tiny-ImageNet) across six Convolutional Neural Networks (CNNs) of different sizes (SqueezeNet 1.1, WideResNet-40-2, WideResNet-16-8, ResNet-18, ResNet-34, ResNet-50). Notably, a single UNIFIED ResNet-34 trained on these datasets achieves accuracies of \textbf{99.53\%}, \textbf{94.84\%}, \textbf{96.12\%}, \textbf{80.32\%}, and \textbf{67.67\%}, respectively, outperforming the average performance of the corresponding independently trained task-specific models while reducing the memory footprint by approximately \textbf{4.94×}, from \textbf{428.30 MB} (five separate models) to \textbf{86.71 MB} (one unified model). UNIFY aims not only to consolidate (K) models into a single model but also to enhance model-effectiveness by leveraging complementary knowledge across datasets. It paves the way for scalable, efficient, and unified learning capable of handling diverse image classification tasks. Code is available at: \href{https://github.com/saktx/UNIFY}{\textbf{https://github.com/saktx/UNIFY}}.
\end{abstract}
\vspace{0.15cm}
\begin{IEEEkeywords}
unified learning, parameter sharing, batch norm switching, memory-efficient learning, edgeAI.
\vspace{-0.10cm}
\end{IEEEkeywords}

\section{Introduction}
\label{sec:intro}

Deep neural networks have achieved remarkable performance across a wide range of computer vision tasks, including image classification, object detection, and image segmentation \cite{lecun2015deep, krizhevsky2012imagenet}. Despite this success, the conventional deep learning paradigm typically trains and maintains a separate model for each dataset or task. This \textbf{“one model per task”} paradigm becomes increasingly inefficient as the number of datasets or tasks grows, limiting the practicality of deploying networks in real-world applications that must operate across multiple domains, such as medical imaging, autonomous driving, agricultural automation, security, and surveillance. For example, an application requiring image classification across five datasets would traditionally use five separate models. In contrast, a unified model could share most of its parameters while retaining only the information necessary to accommodate the dataset-specific differences.

Advances in transfer learning and domain adaptation have addressed the challenge of generalizing across datasets and domains \cite{pan2009survey, hinton2015distilling, gou2021knowledge,yuan2021reinforced, perez2018film}. Transfer learning typically pre-trains a model on a large dataset (e.g. ImageNet-1K) and subsequently fine-tunes it on individual target datasets, leveraging the broader knowledge acquired from the source data. Domain adaptation seeks to mitigate distribution shifts between source and target domains, generally under the assumption that the underlying tasks are closely related \cite{li2016revisiting, he2019rethinking, kornblith2019better, chang2019domain}. Although effective, these approaches generally require separate fine-tuning or adaptation for each target dataset and therefore do not fully exploit the potential of a single, \enquote{one time trained} model to simultaneously handle multiple datasets. This leads to a fundamental question: \textbf{Can a single trained model perform effectively across multiple datasets, while maintaining a similar memory footprint?} HYDRA \cite{sehwag2020hydra} demonstrated that deep networks can provide strong performance even after going through more than 90\% sparsification (99\% for low complexity datasets), revealing that neural networks have huge \enquote{unused} model capacity, motivating a key question: \textbf{Can the unused capacity of a model be leveraged to learn additional datasets?} Inspired by HYDRA, we propose UNIFY, a framework that jointly trains a single model on multiple image-classification datasets, effectively \textbf{“packing”} diverse datasets into its learnable capacity.

\begin{figure*}[!t]
\centerline{\includegraphics[width=1.0\linewidth]{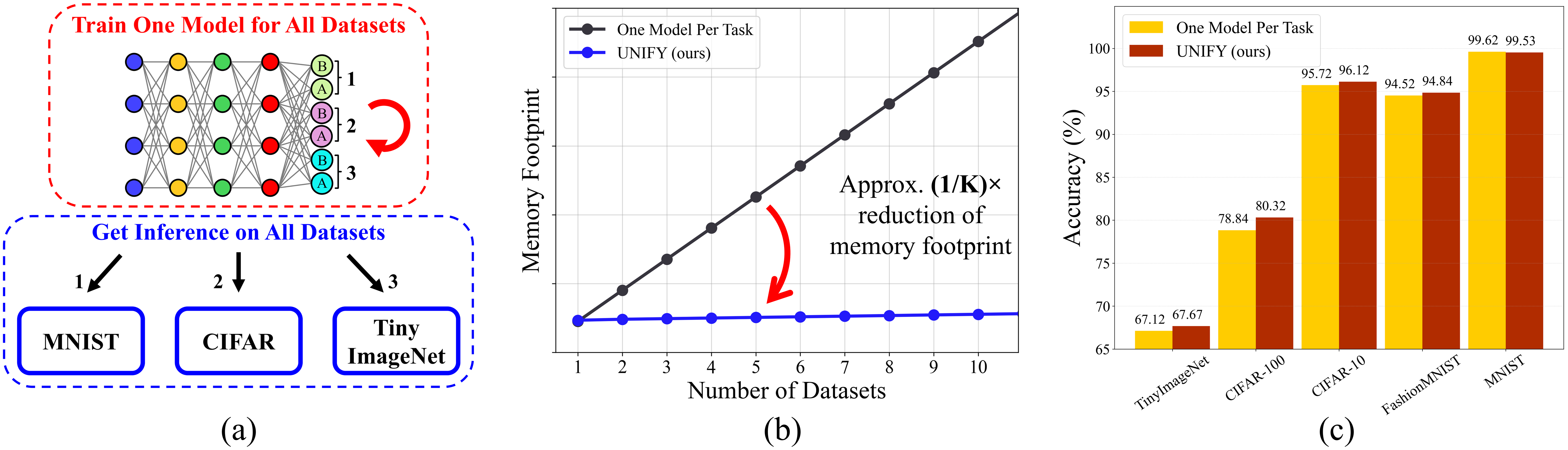}}
\caption{\centering (a) A UNIFIED model jointly trained across multiple datasets enables effective inference on all datasets through Batch Normalization Switching (BNS) and dataset-specific classification heads. (b) A UNIFIED model trained on $K$ datasets, reduces the memory footprint to approximately $1/K$ of that required by $K$ independently trained models, while preserving strong performance. (c) A UNIFIED ResNet-34, jointly trained on 5 datasets, achieves overall superior performance compared with independently trained ResNet-34 models on these datasets.}
\label{fig_1}
\end{figure*}

A major challenge in unified learning stems from substantial differences in input characteristics and feature distributions across datasets, which limits the effectiveness of direct parameter sharing and hinders joint optimization. For example, as shown in Table~\ref{tab:table1} MNIST consists of handwritten digits, whereas Fashion-MNIST contains grayscale images of clothing items. CIFAR-10 and CIFAR-100 contain RGB natural images but differ substantially in their semantic label spaces, while Tiny-ImageNet contains a larger and more diverse collection of natural images. Directly training a single model across such heterogeneous datasets substantially impairs its ability to capture dataset-specific characteristics. These differences necessitate a mechanism that preserves \enquote{dataset-specific} information without requiring duplication of entire model.

UNIFY addresses this problem by using Batch Norm Switching (BNS) and dataset-specific classification heads. BN layers normalize intermediate activations using statistics that capture dataset-specific feature distributions, while dedicated classification heads accommodate differences in the output spaces e.g. CIFAR-10, CIFAR-100, and Tiny-ImageNet require 10, 100, and 200 output neurons, respectively. Rather than duplicating entire models, UNIFY shares the computationally expensive backbone while maintaining separate sets of BN layers and classification heads. During training and inference, a dataset identifier activates the corresponding BN layers and classification head, allowing the shared backbone to process multiple datasets with specialized performance. These switchable components (BN layers and classification heads) constitute a small fraction of the overall model compared with duplicating the entire backbone for each dataset.

\vspace{0.25cm}
\textbf{Key Contributions:} Our contributions are as follows:

\begin{itemize}
    \item We introduce UNIFY, a framework to jointly train one model across 5 datasets, that provides comparable or superior performance compared to independently trained, dataset-specific models.

    \item We employ \textit{Batch Normalization Switching (BNS)} together with \textit{dataset-specific classification heads} that enable multi-dataset (UNIFIED) learning with a single model.

    \item We validate UNIFY using five benchmark datasets (MNIST, Fashion-MNIST, CIFAR-10, CIFAR-100, Tiny-ImageNet) across six different CNN architectures (SqueezeNet 1.1, WideResNet-40-2, WideResNet-16-8, ResNet-18, ResNet-34, ResNet-50), demonstrating its effectiveness and generalization.
    
    \item A single UNIFIED ResNet-34 jointly trained on these datasets achieves accuracies of \textbf{99.53\%}, \textbf{94.84\%}, \textbf{96.12\%}, \textbf{80.32\%}, and \textbf{67.67\%}, respectively, outperforming the average performance of the independently trained ResNet-34 models, while reducing the memory footprint by approximately \textbf{4.94×}, from \textbf{428.30 MB} (five separate models) to \textbf{86.71 MB} (one unified model).  
\end{itemize}

UNIFY is an \textbf{\enquote{architecture-flexible}} method that is not restricted to a single architecture. It can be applied to a variety of CNNs by incorporating Batch Normalization Switching and dataset-specific output heads in their architectures, without fundamentally modifying their feature extraction backbones and without requiring entirely new, specialized architectures. To the best of our knowledge, UNIFY is the first framework to leverage BNS and dataset-specific output heads for joint multi-dataset learning, enabling multiple dataset-specific models to be consolidated into a single UNIFIED model.

\begin{figure*}[t]
\centerline{\includegraphics[width=0.9\linewidth]{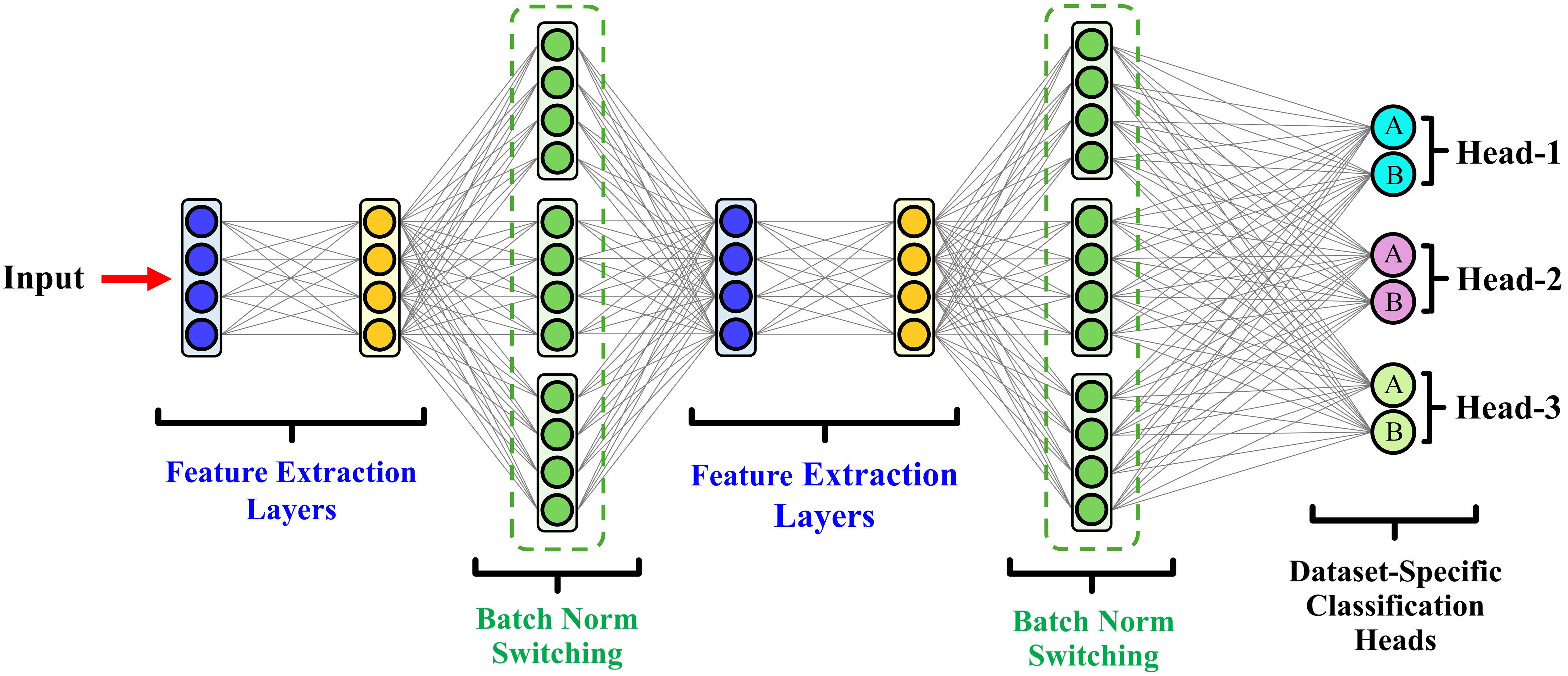}}
\caption{\centering An example UNIFIED architecture for 3 datasets, having 4 feature-extraction layers, 2 switchable BN layers, and an output layer. Each switchable BN layer maintains 3 sets of BN parameters (and statistics), while the output layer contains 3 classification heads, corresponding to the 3 participating datasets. To preserve dataset-specific parameters, the corresponding BN layers and classification head are switched while routing the data from different datasets. The shared feature-extraction layers can be linear, convolutional, or residual connections, depending on the architecture used (UNIFY is architecture-flexible framework that can be adopted in various CNN architectures). Batch Normalization Switching (BNS) and dedicated classification heads preserve dataset-specific representations while enabling joint optimization through a shared backbone.}
\label{fig_2}
\end{figure*}

\section{Related Work}
\label{sec:related_background}
\subsection{Multi-Task Learning}

Multi-Task Learning (MTL) enables multiple related tasks to be learned from shared parameters, promoting knowledge transfer, improving generalization, reducing model redundancy, and enhancing deployment efficiency \cite{standley2020tasks, vandenhende2021multi, crawshaw2020multi, yu2020gradient, sener2018multi, chen2018gradnorm}. The seminal work on MTL \cite{caruana1997multitask} showed that MTL improves generalization due to knowledge sharing. Subsequent work on MTL investigated how task-specific and shared components can be combined and how competing task objectives can be balanced during joint optimization \cite{sener2018multi,yu2020gradient}. Kendall \cite{kendall2018multi} proposed uncertainty-based loss weighting to balance multiple classification and regression objectives and demonstrated that a single model can jointly learn multiple vision tasks. UNIFY focuses on multiple heterogeneous datasets with different visual characteristics, resolutions, distributions, and numbers of classes using a joint objective function.

One model to learn them all \cite{kaiser2017one} designed a \enquote{specialized} architecture, jointly trained across different tasks and domains, using convolutional, attention, and sparsely gated components to accommodate the diverse requirements of different tasks. YOLOR \cite{wang2021you} proposed a unifying framework that integrates explicit and implicit knowledge to enable multi-task learning through kernel-space alignment and prediction refinement, primarily focusing object detection. UNIFY, in contrast, is an \textbf{\enquote{architecture-flexible}} framework that focuses only on image classification and can be adopted in CNNs by means of BNS and dataset-specific output heads, without changing the backbone e.g. a UNIFIED ResNet-34 remains fundamentally a ResNet-34 with switchable BN layers and classification heads.

\subsection{Batch Normalization and Domain-Specific Statistics}

Batch Normalization (BN) was introduced to normalize intermediate activations in a model using “mini-batch statistics” and learnable affine parameters, substantially improving its optimization and training stability \cite{ioffe2015batch}. The learnable parameters and running statistics of a BN layer are associated with the data used for training \cite{ioffe2015batch, bjorck2018understanding}. Consequently, when the trained model is used for a different dataset or domain, the same normalization parameters and statistics cannot adequately capture its feature distribution \cite{szegedy_intriguing_2014, ding_sensitivity_2019, galloway_batch_2019, wang2022removing}. Li et al. proposed Adaptive Batch Normalization (AdaBN) \cite{li2018adaptive}, demonstrating that modifying BN statistics can provide an effective mechanism for domain adaptation without introducing additional trainable components. Similarly, Domain-Specific Batch Normalization (DSBN) \cite{chang2019domain}, demonstrated that domain-specific BN layers can separate domain-specific information from shared representations and can also be extended to multiple source domains.

Although UNIFY and DSBN both maintain separate sets of BN layers, they differ in their objectives and learning settings. DSBN primarily targeted \enquote{unsupervised} domain adaptation across related domains with a shared label space, whereas UNIFY enables supervised joint learning across heterogeneous datasets with varying characteristics and class counts. UNIFY further incorporates dataset-specific classification heads alongside switchable BN layers, jointly optimizing these components and the shared backbone using cross-entropy losses. It aims at consolidating $K$ dataset-specific models into a single UNIFIED model, to reduce deployment memory to approximately $1/K$ of the original requirement, with minimal parameter overhead, while maintaining model performance.

\subsection{Normalization Switching}
Switchable Normalization (SN) \cite{luo2018differentiable}, learns to combine Instance Normalization, Layer Normalization, and Batch Normalization and their associated statistics, demonstrating that normalization can be adapted according to the network layer and learning problem. Similarly, OAT \cite{wang2020once} used Dual Batch Normalization for distinguishing clean and adversarial data and used switchable DBN in OATS (architecture slimming). Task-Specific Normalization \cite{zhang2024task} \enquote{freezes} the network backbone and learns task-specific normalization parameters using a separate prediction head for each task. These studies show that BN can serve not only as an optimization mechanism but also as a lightweight component for accommodating distributional differences in the data.
UNIFY, on the contrary, explicitly performs dataset-dependent BNS and treats normalization as a mechanism for specializing different datasets using a UNIFIED model, without freezing the backbone, conceptually positioned at the intersection of shared-representation learning, multi-task learning, and data-specific normalization.

\section{UNIFY -- Methodology}
\subsection{Overview}
UNIFY is designed around a simple idea: \textit{share the parameter-intensive backbone and specialize through lightweight components}. A single UNIFIED model comprises three primary components:
\vspace{0.2cm}
\begin{enumerate}
    \item A shared backbone.
    \item Set of dataset-specific Batch Normalization (BN) layers.
    \item Set of dataset-specific classification heads.
\end{enumerate}
\vspace{0.2cm}
Let the set of selected datasets be denoted as:
\begin{equation}
\mathcal{D} = \{\mathcal{D}_1, \mathcal{D}_2, \ldots, \mathcal{D}_K\}
\end{equation}

A UNIFIED network $\mathcal{N}$ consists of a shared backbone $\mathcal{F}$ with dataset-specific sets of BN layers $\mathcal{B}_k$ and separate classification heads $\mathcal{C}_k$:

\begin{equation}
\mathcal{B} = \{\mathcal{B}_1, \mathcal{B}_2, \ldots, \mathcal{B}_K\}
\end{equation}

\begin{equation}
\mathcal{C} = \{\mathcal{C}_1, \mathcal{C}_2, \ldots, \mathcal{C}_K\}
\end{equation}

For an input $x$ originating from dataset $\mathcal{D}_k$, the same shared backbone $\mathcal{F}(\cdot;\theta)$ is used, while the set of BN layers $\mathcal{B}_k$ and the dataset-specific classification head $\mathcal{C}_k(\cdot;\phi_k)$ (associated with $\mathcal{D}_k$ in the architecture) are activated to process the input through the dataset-specific pathway. The overall prediction can be expressed as:
\begin{equation}
\hat{y} =
\mathcal{C}_k\left(\mathcal{F}(x;\theta,\mathcal{B}_k);\phi_k\right)
\end{equation}
where $\theta$ represents the shared backbone parameters, $\mathcal{B}_k$ represents the BN parameters and statistics associated with dataset $\mathcal{D}_k$, and $\phi_k$ represents the dataset-specific parameters of the classification head $\mathcal{C}_k$.

\begin{figure}[t]
    \centering
    \includegraphics[width=1.0\linewidth]{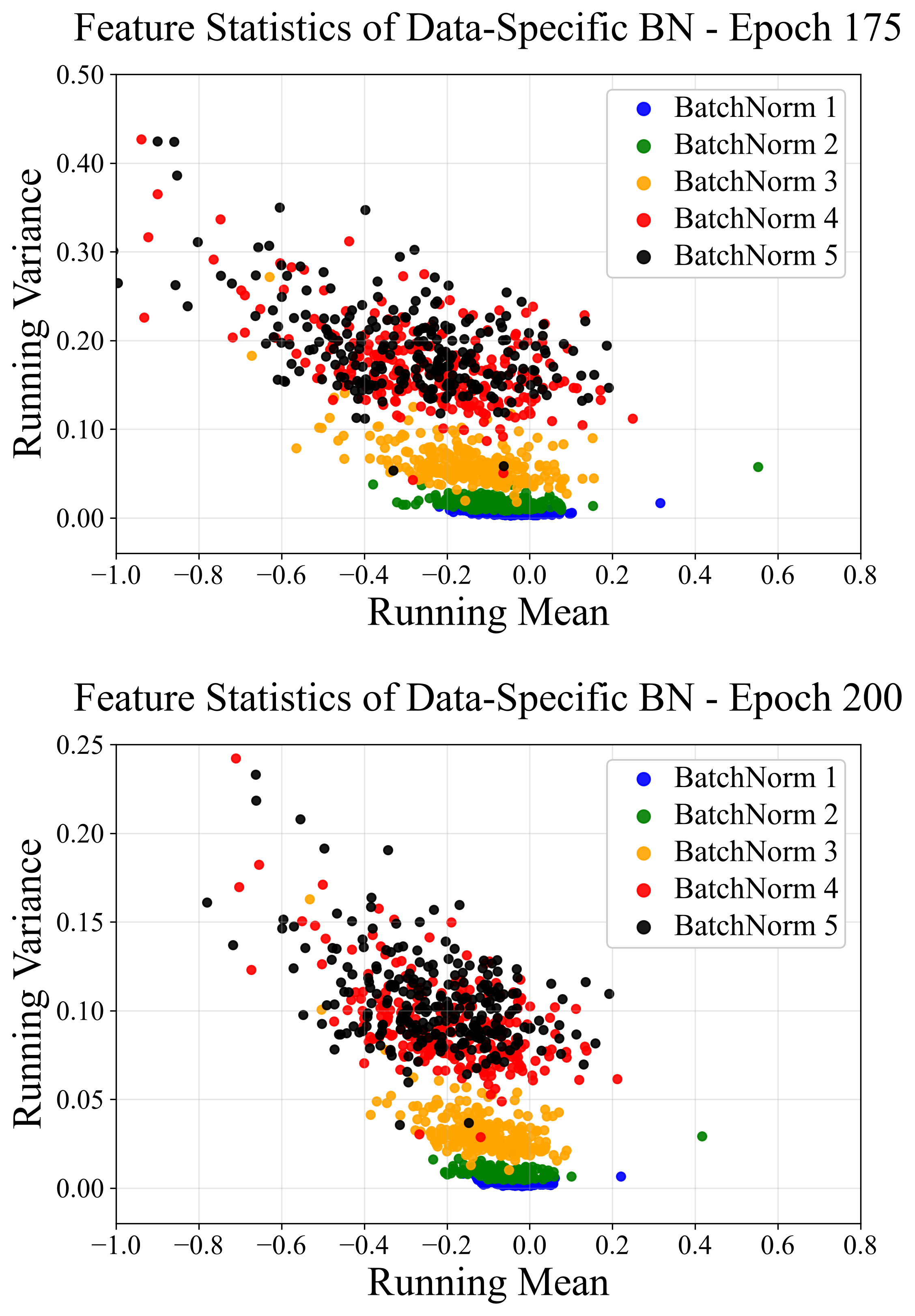}
    \caption{\centering Visualization of the feature statistics in a Switchable BN layer of a UNIFIED SqueezeNet 1.1 during training. BN Heads 1 to 5 correspond to MNIST, Fashion-MNIST, CIFAR-10, CIFAR-100, and Tiny-ImageNet, respectively. The differences in the statistical distributions across the 5 BN sets highlight the importance of BNS in accommodating dataset-specific feature characteristics. These results suggest that sharing a single BN configuration is ineffective and sub-optimal for accommodating distinct data characteristics of the 5 datasets, resulting in degraded performance, as also supported by the results in Table~\ref{tab:table4}.}
    \label{fig:figure3}
\end{figure}

\subsection{Shared Backbone}
The backbone is responsible for learning feature representations across the participating datasets. UNIFY shares the memory-intensive backbone across all datasets:
\begin{equation}
\theta_1 = \theta_2 = \cdots = \theta_K = \theta
\end{equation}
Only the BN configurations and classification heads are dataset-specific. The conventional multi-dataset learning trains independent and separate models, whose feature-extraction layers are replicated for every dataset e.g. when $K$ datasets are considered, independent training produces $K$ sets of entire model parameters:
\begin{equation}
\Theta_1,\Theta_2,\Theta_3,\ldots,\Theta_K\
\end{equation}

In contrast, UNIFY maintains a single shared set of backbone parameters and augments it only with dataset-specific BN layers and classification heads:
\begin{equation}
\Theta_{\mathrm{UNIFY}}
=
\left\{
\theta,
\mathcal{B}_1,\ldots,\mathcal{B}_K,
\phi_1,\ldots,\phi_K
\right\}
\end{equation}

Therefore, the parameter-intensive backbone (that has high memory footprint) is instantiated only once and is jointly optimized using training signals from the participating datasets.

\subsection{Batch Normalization Switching (BNS)}

Batch Normalization Switching (BNS) is the main mechanism that enables dataset-specific adaptation in UNIFY. For an activation $x$, a BN layer can be expressed as:
\begin{equation}
\hat{x} =
\frac{x-\mu}{\sqrt{\sigma^2+\epsilon}}
\end{equation}
\begin{equation}
y = \gamma\hat{x}+\beta
\end{equation}
where $\mu$ and $\sigma^2$ denote the mean and variance, respectively, $\epsilon$ is a small constant added for numerical stability, and $\gamma$ and $\beta$ are the learnable scale and shift parameters.

A conventional model (such as WideResNet-16-8 or ResNet-34) uses a single set of BN layers in its architecture, maintaining only one set of running statistics and affine parameters. Sharing these BN parameters across multiple datasets substantially degrades its performance across these datasets, since a common set of BN layers cannot adequately capture the distinct feature distributions of different datasets. UNIFY assigns each dataset its own BN configuration:
\begin{equation}
\mathcal{B}_k =
\{\mu_k,\sigma_k^2,\gamma_k,\beta_k\}
\end{equation}

For an input originating from dataset $\mathcal{D}_k$, the corresponding configuration $\mathcal{B}_k$ is selected at both training and inference time. The backbone remains shared while the normalization operation adapts to the statistics of the corresponding dataset e.g. when a UNIFIED ResNet-18 processes a Fashion-MNIST image followed by a CIFAR-100 image, the backbone remains unchanged but the BN configuration switches according to the dataset:
\begin{equation}
\mathcal{D}_{\mathrm{Fashion-MNIST}} \rightarrow \mathcal{B}_{\mathrm{Fashion-MNIST}}
\end{equation}

\begin{equation}
\mathcal{D}_{\mathrm{CIFAR-100}} \rightarrow \mathcal{B}_{\mathrm{CIFAR-100}}
\end{equation}

BNS is important because different datasets generally exhibit different data distributions (see Fig.~\ref{fig:figure3}). The feature distributions of handwritten digit images differ considerably from those of natural RGB images.

\subsection{Dataset-Specific Classification Heads}

Although the backbone can be shared, the output spaces of the participating datasets may not be necessarily identical (see Table~\ref{tab:table1}). Even if multiple datasets have identical number of classes, a shared classification head cannot adequately capture their dataset-specific output distributions as discussed in \cite{tareen2026solar}. UNIFY therefore employs a separate classification head for each dataset. Let $z$ denote the shared feature representation for an input $x$ from dataset $\mathcal{D}_k$:

\begin{equation}
z = \mathcal{F}(x;\theta,\mathcal{B}_k)
\end{equation}

The corresponding output head $\mathcal{C}_k$ provides the prediction:
\begin{equation}
\hat{y}_k = \mathcal{C}_k(z;\phi_k)
\end{equation}

Therefore, when a Fashion-MNIST sample is processed, the set of Fashion-MNIST-specific BN layers along with its dedicated classification head are activated, whereas a CIFAR-10 sample is routed through the CIFAR-10-specific BN layers and output head. This multi-head design prevents incompatible output spaces from being forced into a single classifier.

\begin{table}[t]
    \centering
    \caption{Details of the datasets used in our experiments.}
    \label{tab:table1}
    \resizebox{\columnwidth}{!}{%
    \begin{tabular}{lclc}
        \hline
        \textbf{Dataset} & \textbf{Image Size} & \textbf{Description} & \textbf{Classes} \\
        \hline
        MNIST & $28\times28\times1$ & Handwritten digits & 10 \\
        Fashion-MNIST & $28\times28\times1$ & Clothing and apparel & 10 \\
        CIFAR-10 & $32\times32\times3$ & Natural images & 10 \\
        CIFAR-100 & $32\times32\times3$ & Fine-grained natural images & 100 \\
        Tiny-ImageNet & $64\times64\times3$ & Diverse natural images & 200 \\
        \hline
    \end{tabular}%
    }
\end{table}

\subsection{Joint Training}

Algorithm~\ref{alg:algo1} shows the joint training process of UNIFY. At each training iteration, UNIFY processes a mini-batch of data from each dataset. For a batch $(x,y)$ sampled from dataset $\mathcal{D}_k$, the corresponding BN configuration $\mathcal{B}_k$ and classification head $\mathcal{C}_k$ are activated. The input is then propagated through the shared backbone and the dataset-specific components. The prediction can be expressed as:

\begin{equation}
\hat{y}_k =
\mathcal{C}_k\left(
\mathcal{F}(x;\theta,\mathcal{B}_k);
\phi_k
\right)
\end{equation}

where $\theta$ denotes the shared backbone parameters and $\phi_k$ denotes the parameters of the $k$-th classification head. The corresponding dataset-specific loss is given by:

\begin{equation}
\mathcal{L}_k =
\mathcal{L}_{\mathrm{CE}}(\hat{y}_k,y)
\end{equation}

where $\mathcal{L}_{\mathrm{CE}}$ denotes the cross-entropy loss. The losses obtained from all $K$ datasets are accumulated to form a joint training objective:

\begin{equation}
\mathcal{L} = \mathcal{L}_{1} + \mathcal{L}_{2} + \cdots + \mathcal{L}_{k}
\end{equation}

Gradients of the joint loss are subsequently backpropagated through the UNIFIED network and the optimizer updates the shared and dataset-specific parameters accordingly. The shared backbone receives optimization signals from all participating datasets, while the dataset-specific components are specialized according to the dataset-specific features. By leveraging complementary knowledge across datasets, joint training promotes richer shared representations and leads to improved generalization and overall performance for some models.

\subsection{Inference}

During inference, UNIFY selects the dataset-specific BN configuration and classification head to process inputs from the target dataset. A dataset identifier acts as the routing signal that determines the appropriate BN configuration and classification head. Each dataset has its own inference pathway through the data-specific components while the backbone parameters $\theta$ are shared. Consequently, a single model can be deployed while supporting inference across multiple datasets.

\begin{algorithm}[t]
\caption{Training a UNIFIED Model}
\label{alg:algo1}
\footnotesize
\begin{algorithmic}[1]  
\Require Set of training datasets $\mathcal{D}$ = \{$\mathcal{D}_{1}$, $\mathcal{D}_{2}$, ..., $\mathcal{D}_{K}$\}, UNIFY Network $\mathcal{N}$, max iterations $T$
\Ensure Trained network parameters $\Theta_{\mathrm{UNIFY}}$
\State Initialize parameters $\Theta_{\mathrm{UNIFY}}$ of $\mathcal{N}$
\State Initialize $optimizer$ with $\Theta_{\mathrm{UNIFY}}$
\For{$t = 1$ to $T$}
    \State Clear gradients: $optimizer.zero\_grad()$
    \For{dataset $\mathcal{D}_{k} \in \mathcal{D}$}
    	\State Sample batch $(x, y)$ from $\mathcal{D}_{k}$
        \State Activate $\mathcal{B}_{k}$ in $\mathcal{N}$
        \State Activate classification head $\mathcal{C}_k$ in $\mathcal{N}$

        \State Compute dataset-specific loss $\mathcal{L}_{k}(x, y)$
    \EndFor
    \State Calculate total loss: $\mathcal{L} = \mathcal{L}_{1} + \mathcal{L}_{2} + ...$ $\mathcal{L}_{k}$
    \State Backpropagate: $\mathcal{L}.backward()$
    \State Update parameters: $optimizer.step()$
\EndFor
\State \Return $\Theta_{\mathrm{UNIFY}}$
\end{algorithmic}
\end{algorithm}

\subsection{Parameter and Memory Efficiency}

UNIFY is efficient because it avoids duplicating the entire network backbone and uses light-weight components (BN layers and classification heads) that constitute a small fraction of the original model e.g. BN layers and a classification head in a ResNet-34 has less than 0.5\% parameters as compared to the full model. For $K$ independently trained models, the aggregate parameter requirement is expressed as:
\begin{equation}
\mathcal{P}_{\mathrm{Separate}}
=
\sum_{k=1}^{K}\mathcal{P}_k
\end{equation}
where $\mathcal{P}$ denotes the parameter count of a single model. When the same architecture is independently trained for $K$ datasets, the aggregate parameters become:
\begin{equation}
\mathcal{P}_{\mathrm{Separate}} \approx K\mathcal{P}
\end{equation}

In contrast, UNIFY requires only one shared backbone with dataset-specific sets of BN layers and classification heads:
\begin{equation}
\mathcal{P}_{\mathrm{UNIFIED}}
=
\mathcal{P}_{\mathrm{Backbone}}
+
\sum_{k=1}^{K}
\left(
P_{\mathrm{BN},k}
+
P_{\mathrm{Head},k}
\right)
\end{equation}

Since the BN layers and output heads introduce minimal additional parameters, the memory footprint of the UNIFIED model is substantially smaller than the footprint of $K$ independently trained models.

\section{Experiments, Results, \& Discussion}

\begin{table*}[t]
    \centering
    \caption{\centering Performance comparison of independently trained dataset-specific models with corresponding UNIFIED models jointly trained across all datasets (\textcolor{my_blue}{\textbf{bold blue}} and \textbf{bold black} indicate superior performance achieved by UNIFIED and dataset-specific models, respectively). Results demonstrate that a single UNIFIED model can provide comparable or even superior performance on all datasets as compared to independently trained models.}
    \label{tab:table2}
    
    \setlength{\tabcolsep}{6pt}
    \renewcommand{\arraystretch}{0.75}
    \footnotesize
    
    \begin{tabular}{lccccc}
        \toprule
        \textbf{Model} &
        \textbf{MNIST} &
        \textbf{Fashion-MNIST} &
        \textbf{CIFAR-10} &
        \textbf{CIFAR-100} &
        \textbf{Tiny-ImageNet} \\
        \midrule
        
        
        
        
        SqueezeNet 1.1
        & 99.39 & 92.18 & 89.42 & 66.24 & \textbf{57.06} \\
        
        SqueezeNet 1.1 (\tiny{UNIFIED})
        & \textcolor{my_blue}{\textbf{99.41}} & \textcolor{my_blue}{\textbf{93.12}} & \textcolor{my_blue}{\textbf{91.06}} & \textcolor{my_blue}{\textbf{69.18}} & 56.76 \\
        
        \addlinespace
        
        WideResNet-40-2
        & 99.61 & \textbf{94.36} & \textbf{94.91} & \textbf{76.71} & 60.98 \\
        
        WideResNet-40-2 (\tiny{UNIFIED})
        & \textcolor{my_blue}{\textbf{99.62}} & 94.22 & 94.39 & 75.29
        & \textcolor{my_blue}{\textbf{61.13}} \\
        
        \addlinespace
        
        
        
        
        WideResNet-16-8
        & 99.60 & \textbf{94.55} & 95.73 & \textbf{80.22} & \textbf{65.81} \\
        
        WideResNet-16-8 (\tiny{UNIFIED})
        & \textcolor{my_blue}{\textbf{99.66}} & 94.44
        & \textcolor{my_blue}{\textbf{96.00}} & 79.41 & 65.19 \\
        
        \addlinespace
        
        ResNet-18
        & \textbf{99.61} & 94.50 & 95.40 & 77.84 & 65.87 \\
        
        ResNet-18 (\tiny{UNIFIED})
        & 99.56 & \textcolor{my_blue}{\textbf{94.84}}
        & \textcolor{my_blue}{\textbf{95.44}}
        & \textcolor{my_blue}{\textbf{78.21}}
        & \textcolor{my_blue}{\textbf{66.22}} \\
        
        \addlinespace
        
        ResNet-34
        & \textbf{99.62} & 94.52 & 95.72 & 78.84 & 67.12 \\
        
        ResNet-34 (\tiny{UNIFIED})
        & 99.53 & \textcolor{my_blue}{\textbf{94.84}}
        & \textcolor{my_blue}{\textbf{96.12}}
        & \textcolor{my_blue}{\textbf{80.32}}
        & \textcolor{my_blue}{\textbf{67.67}} \\
        
        \addlinespace
        
        ResNet-50
        & 99.60 & 94.17 & 95.49 & 78.57 & 68.81 \\
        
        ResNet-50 (\tiny{UNIFIED})
        & \textcolor{my_blue}{\textbf{99.63}}
        & \textcolor{my_blue}{\textbf{94.48}}
        & \textcolor{my_blue}{\textbf{95.64}}
        & \textcolor{my_blue}{\textbf{80.22}}
        & \textcolor{my_blue}{\textbf{69.67}} \\
        
        \bottomrule
    \end{tabular}
\end{table*}

\begin{table*}[htbp]
    \centering
    \caption{\centering Comparison of parameter count and memory footprint between dataset-specific and UNIFIED models at Float32 precision. The percentage parameter and memory overhead introduced by the dataset-specific BN layers and classification heads is reported in parentheses. The deployment-size reduction shows the reduction achieved by replacing 5 independently deployed models with 1 UNIFIED model.}
    \label{tab:table3}

    \setlength{\tabcolsep}{7pt}
    \renewcommand{\arraystretch}{0.75}
    \footnotesize

    \begin{tabular}{lllc c}
        \toprule
        \textbf{Model} &
        \textbf{Parameters} &
        \textbf{Memory Footprint} &
        \textbf{Deployment-Size Reduction} \\
        \midrule



        
        SqueezeNet 1.1
        & 0.82 M & 3.14 MB & -- \\

        SqueezeNet 1.1 (\tiny{UNIFIED})
        & 0.92 M ($\uparrow$ 12.19\%) & 3.82 MB ($\uparrow$ 21.66\%)
        & $\downarrow 4.11\times$ \\

        \addlinespace
        
        WideResNet-40-2
        & 2.26 M & 9.17 MB & -- \\

        WideResNet-40-2 (\tiny{UNIFIED})
        & 2.31 M ($\uparrow$ 2.21\%) & 9.68 MB ($\uparrow$ 5.56\%)
        & $\downarrow 4.74\times$ \\

        \addlinespace



        
        WideResNet-16-8
        & 11.05 M & 44.29 MB & -- \\

        WideResNet-16-8 (\tiny{UNIFIED})
        & 11.15 M ($\uparrow$ 0.90\%) & 44.88 MB ($\uparrow$ 1.33\%)
        & $\downarrow 4.93\times$ \\

        \addlinespace

        ResNet-18
        & 11.27 M & 45.16 MB & -- \\

        ResNet-18 (\tiny{UNIFIED})
        & 11.37 M ($\uparrow$ 0.89\%) & 45.87 MB ($\uparrow$ 1.57\%)
        & $\downarrow 4.92\times$ \\

        \addlinespace

        ResNet-34
        & 21.38 M & 85.66 MB & -- \\

        ResNet-34 (\tiny{UNIFIED})
        & 21.51 M ($\uparrow$ 0.61\%) & 86.71 MB ($\uparrow$ 1.23\%)
        & $\downarrow 4.94\times$ \\

        \addlinespace

        ResNet-50
        & 23.91 M & 95.96 MB & -- \\

        ResNet-50 (\tiny{UNIFIED})
        & 24.39 M ($\uparrow$ 2.01\%) & 99.11 MB ($\uparrow$ 3.28\%)
        & $\downarrow 4.84\times$ \\

        \bottomrule
    \end{tabular}
\end{table*}

\begin{table*}[t]
    \centering
    \caption{\centering Performance of UNIFIED models without the use of Batch Normalization Switching (BNS). After 200 epochs of training, the results demonstrate that a common BN configuration does not adequately accommodate the distinct feature distributions of the 5 datasets even if dedicated classification heads are used, leading to overall degraded performance.}
    \label{tab:table4}

    \setlength{\tabcolsep}{4pt}
    \renewcommand{\arraystretch}{0.75}
    \footnotesize

    \begin{tabular}{lccccc}
        \toprule
        \textbf{Model} &
        \textbf{MNIST} &
        \textbf{Fashion-MNIST} &
        \textbf{CIFAR-10} &
        \textbf{CIFAR-100} &
        \textbf{Tiny-ImageNet} \\
        \midrule


		
        SqueezeNet 1.1
        & 89.68 & 72.10 & 85.42 & 59.46 & 33.26 \\

        \addlinespace		
		
        WideResNet-40-2
        & 70.08 & 78.34 & 87.48 & 64.09 & 33.06 \\

        \addlinespace

        WideResNet-16-8
        & 88.89 & 84.39 & 86.98 & 54.92 & 26.23 \\

        \addlinespace

        ResNet-18
        & 91.46 & 87.31 & 91.13 & 66.65 & 47.28 \\

        \addlinespace

        ResNet-34
        & 91.60 & 87.37 & 92.34 & 71.40 & 46.19 \\

        \addlinespace

        ResNet-50
        & 86.45 & 84.83 & 87.01 & 60.04 & 37.84 \\

        \bottomrule
    \end{tabular}
\end{table*}

UNIFY is validated using six CNN architectures of different sizes (SqueezeNet 1.1, WideResNet-40-2, WideResNet-16-8, ResNet-18, ResNet-34, ResNet-50) across five datasets (MNIST, Fashion-MNIST, CIFAR-10, CIFAR-100, Tiny-ImageNet). For all experiments, UNIFIED models were jointly trained across five datasets for $200$ epochs with a batch size of $64$. The dataloaders of smaller datasets (MNIST, Fashion-MNIST, CIFAR-10, CIFAR-100) were repeated to match the length of the largest dataset (Tiny-ImageNet). Training employed SGD optimizer with an initial learning rate of $0.1$, momentum of $0.9$, and weight decay of $5\times10^{-4}$, using a cosine annealing learning rate schedule. Same training settings were used to train independent dataset-specific models to ensure a fair comparison.

Table~\ref{tab:table2} presents a comparison of UNIFIED models (jointly trained across five datasets) with the independently trained dataset-specific models as baselines. Overall, UNIFY demonstrates competitive or superior performance, outperforming the baseline models in 21 out of 30 architecture--dataset combinations, despite consolidating five dataset-specific models into single UNIFIED model for each architecture. Notably, the UNIFIED SqueezeNet 1.1 achieves higher accuracy on four of the five datasets, with improvements of 1.64 and 2.94 percentage points on CIFAR-10 and CIFAR-100, respectively. Similarly, UNIFIED ResNet-18 and ResNet-34 outperform their dataset-specific counterparts on four datasets each, with UNIFIED ResNet-34 achieving improvements of 0.40, 1.48, and 0.55 percentage points on CIFAR-10, CIFAR-100, and Tiny-ImageNet, respectively. The UNIFIED ResNet-50 demonstrates particularly consistent improvements, outperforming its independently trained counterparts across all five datasets, including gains of 1.65 and 0.86 percentage points on CIFAR-100 and Tiny-ImageNet, respectively. In contrast, the UNIFIED WideResNet architectures exhibit modest performance reductions on several datasets, indicating that the benefits of unified learning can vary with CNN architecture. Nevertheless, these results demonstrate that UNIFY can effectively consolidate multiple dataset-specific models into a single UNIFIED model while maintaining comparable or even superior classification accuracy. 

Table~\ref{tab:table3} further demonstrates the main advantage of UNIFY in terms of memory-efficiency. Except SqueezeNet 1.1 which is a tiny model, introducing dataset-specific BN and classification heads in the other five CNNs increases the parameter count by no more than \textbf{2.21\%} and the memory footprint by no more than \textbf{5.56\%} relative to corresponding dataset-specific models. Yet, a single UNIFIED model replaces five independently deployed models, yielding deployment-size reductions ranging from \textbf{4.11$\times$ to 4.94$\times$}, thereby substantially reducing storage requirements. For example, the UNIFIED ResNet-34 requires only \textbf{21.51 M parameters} with a memory footprint of \textbf{86.71 MB}, compared with \textbf{21.38 M parameters} and \textbf{85.66 MB} for a single independently trained ResNet-34. Nevertheless, the UNIFIED model replaces five such models requiring \textbf{428.30 MB} in total while achieving comparable or superior performance across all five datasets. In the case of UNIFIED SqueezeNet 1.1, the parameter count and memory footprint increase more (i.e. by 12.19\% and 21.66\%, respectively) because it is a small model with only \textbf{0.82 M} parameters.

Table~\ref{tab:table4} presents an important ablation of BNS. When a common BN configuration is used, instead of BNS, across all datasets during unified learning, even with dedicated classification heads, the performance of the models degrades substantially across every architecture and dataset. Table~\ref{tab:table4} shows that, without BN Switching, WideResNet-40-2 achieves accuracies of 70.08\%, 78.34\%, 87.48\%, 64.09\%, and 33.06\%, while ResNet-50 achieves 86.45\%, 84.83\%, 87.01\%, 60.04\%, and 37.84\% on MNIST, Fashion-MNIST, CIFAR-10, CIFAR-100, and Tiny-ImageNet, respectively. Compared with the corresponding UNIFY models, the performance drop is substantially high across all 5 datasets e.g. WideResNet-40-2 exhibits performance drops of 29.54, 15.88, 6.91, 11.20, and 28.08 percentage points, while ResNet-50 exhibits drops of 13.18, 9.65, 8.63, 20.18, and 31.83 percentage points on MNIST, Fashion-MNIST, CIFAR-10, CIFAR-100, and Tiny-ImageNet, respectively. Similar trends can be observed for the other 3 models.

This substantial performance degradation demonstrates that dataset-specific classification heads alone are insufficient to accommodate the heterogeneous feature distributions encountered during joint training (see Fig.~\ref{fig:figure3}). The results validate BNS as a critical component of UNIFY, enabling the shared backbone to retain dataset-specific feature normalization while avoiding the redundancy of maintaining separate complete models.

\section{Future Directions}
We plan to explore the broader applicability of UNIFY across additional datasets, architectures, tasks, and modalities in future. UNIFY will be evaluated on large-scale datasets such as ImageNet, which was beyond the computational resources available in the current study. We also plan to test UNIFY for object detection by integrating BNS into shared detection backbones to examine whether the framework can unify multiple detection datasets while retaining a common backbone. Further evaluations will consider image segmentation, depth estimation, and multi-modal tasks, as well as TinyML models. Other interesting future directions include automatic dataset-aware switching, continual dataset integration, and complementary techniques such as pruning, quantization, and knowledge distillation.

\section*{Conclusion}
We presented \textbf{UNIFY}, an architecture-flexible framework that enables unified learning with one model across multiple heterogeneous datasets via Batch Normalization Switching (BNS). By sharing the parameter-intensive backbones in CNNs and incorporating lightweight dataset-specific BN configurations and classification heads, UNIFY effectively balances parameter sharing and dataset-specific adaptation without requiring separate complete models. Extensive experiments across five benchmark datasets and six different CNN architectures demonstrate that UNIFY achieves comparable or superior performance compared with independently trained dataset-specific models, while substantially reducing model redundancy and deployment memory requirements. UNIFY is a simple, scalable, and memory-efficient framework for constructing UNIFIED models capable of supporting multiple datasets and tasks through a single UNIFIED neural network.

\bibliographystyle{IEEEtran}
\bibliography{main}
\end{document}